%% file: NeuralVideo.tex
\documentclass{article}
\usepackage{times}
\usepackage{graphicx} 
\usepackage{subfigure} 
\usepackage{natbib}
\usepackage{algorithm}
\usepackage{algorithmic}
\usepackage{hyperref}

\usepackage[accepted]{icml2017} 

\hypersetup{
  pdfinfo={
    Title={Memory-augmented Attention Modelling for Videos},
    Author={Rasool Fakoor, Abdel-rahman Mohamed, Margaret Mitchell, Sing Bing Kang, Pushmeet Kohli},
  }
}
\pdfoutput=1

\usepackage{amsmath}
\usepackage{graphicx}
\usepackage{tabularx}
\usepackage{booktabs}
\usepackage{tablefootnote}
\usepackage{mdframed}
 \usepackage{enumitem}
 \usepackage{multirow}

\title{Memory-augmented Attention Modelling for Videos}

\date{}
\author{Rasool Fakoor$^{\dag}$\thanks{ Corresponding author: Rasool Fakoor (
rasool.fakoor@mavs.uta.edu)},
Abdel-rahman Mohamed,$^{\dag\dag}$,
Margaret Mitchell$^{\ddag\dag}$,
Sing Bing Kang$^{\dag\dag}$,\\
Pushmeet Kohli$^{\dag\dag}$\\
$^{\dag\dag}$Microsoft Research ~~
$^{\dag}$University of Texas at Arlington~~
$^{\ddag\dag}$Google\\
$^{\dag}$\small{rasool.fakoor@mavs.uta.edu},
$^{\dag\dag}$\small{\{asamir,~singbing.kang,~pkohli\}@microsoft.com}
$^{\ddag\dag}$\small{mmitchellai@google.com},
}
\usepackage{etoolbox}
\makeatletter
\patchcmd\@combinedblfloats{\box\@outputbox}{\unvbox\@outputbox}{}{%
  \errmessage{\noexpand\@combinedblfloats could not be patched}%
}%
\makeatother

\begin{document}

\maketitle


\input{Abstract}
\input{Introduction}
\input{RelatedWorks}
\input{Model}

\input{Experiments}
\input{Conclusion}

\bibliography{ref}
\bibliographystyle{icml2017}

\end{document}

%% file: Abstract.tex
\begin{abstract}
 
We present a method to improve video description generation by modeling higher-order interactions between video frames and described concepts. By storing past visual attention in the video associated to previously generated words, the system is able to decide what to look at and describe in light of  
what it has already looked at and described. This enables not only more effective local attention, but tractable consideration of the video sequence while generating each word. Evaluation on the challenging and popular {\it MSVD} and {\it Charades} datasets demonstrates that the proposed architecture outperforms previous video description approaches without requiring  external temporal video features.  The source code for this paper is available on \url{https://github.com/rasoolfa/videocap}.

\end{abstract} 

%% file: Introduction.tex
\section{Introduction}

Deep neural architectures have led to remarkable progress in computer vision and natural language processing problems. Image captioning is one such problem, where the combination of convolutional structures~\citep{alexnet, lecun1998gradient}, 
and sequential recurrent structures \citep{s2sIlya} leads to remarkable improvements over previous work \cite{FangEtAl2015,DevlinEtAl2015}.
One of the emerging modelling paradigms,  shared by models for image captioning as well as related vision-language problems, is the notion of an attention mechanism that guides the model to attend to certain parts of the image while generating \cite{icml2015_xuc15}.

The attention models used for problems such as image captioning typically depend on the single image under consideration and the partial output generated so far, jointly capturing one region of an image and the words being generated.  However, such models cannot directly capture the temporal reasoning necessary to effectively produce words that refer to actions and events taking place over multiple frames in a video.  For example, in a video depicting ``someone waving a hand'', the ``waving'' action can start from any frame and can continue on for a variable number of following frames. At the same time, videos contain many frames that do not provide additional information over the smaller set of frames necessary to generate a summarizing description. Given these challenges, it is not surprising that even with recent advancements in image captioning~\cite{FangEtAl2015, icml2015_xuc15, densecap, Vinyals_2015_CVPR, lrcn2014}, video captioning has remained challenging.

Motivated by these observations, we introduce a memory-based attention mechanism for video captioning and description.  Our model utilizes memories of past attention in the video when reasoning about where to attend in a current time step. This allows the model to not only effectively leverage local attention, but also to consider the entire video as it generates each word.  This mechanism effectively binds information from both vision and language sources into a coherent structure.  

Our work shares the same goals as recent work on attention mechanisms for sequence-to-sequence architectures, such as \citet{RocktaschelGHKB15} and \citet{YangYWSC16}.  
~\citet{RocktaschelGHKB15} consider the domain of entailment relations, where the goal is to determine entailment given two input sentences. They propose a soft attention model that is not only focused on the current state, but the previous as well.  In our model, all previous attentions are explicitly stored into memory, and the system learns to memorize the encoded version of the input videos conditioned on previously seen words. \citet{YangYWSC16} and our work both try to solve the problem of locality of attention in vision-to-language, but while \citet{YangYWSC16} introduce a memory architecture optimized for single image caption generation, we introduce a memory architecture that operates on a streaming video's temporal sequence. 

The contributions of this work include:
\begin{itemize}
    \item  A deep learning architecture that represents video with an explicit model of the video's temporal structure. \vspace{-.5em}
    \item A method to jointly model the video description and temporal video sequence, connecting the visual video space and the language description space. \vspace{-.5em}
    \item  A memory-based attention mechanism that learns iterative attention relationships in a simple and effective sequence-to-sequence memory structure. \vspace{-.5em}
    \item Extensive comparison of this work and previous work on the video captioning problem on the MSVD \citep{chencl11} and the Charades \citep{sigurdsson2016hollywood} datasets.  
\end{itemize}

\noindent We focus on the video captioning problem, however, the proposed model is general enough to be applicable in other sequence problems where attention models are used (e.g., machine translation or recognizing entailment relations).

%% file: RelatedWorks.tex
\section{Related Work}

One of the primary challenges in learning a mapping from a visual space (i.e., video or image) to a language space is learning a representation that not only effectively represents each of these modalities, but is also able to translate a representation from one space to the other. \citet{Rohrbachiccv2013} developed a model that generates a semantic representation of visual content that can be used as the source language for the language generation module. \citet{venugopalannaacl15} proposed a deep method to translate a video into a sentence where an entire video is represented with a single vector based on the mean pool of frame features. However, it was recognized that representing a video by an average of its frames loses the temporal structure of the video. To address this problem, recent work \citep{yao2015capgenvid, pan2016hierarchical, venugopalan15iccv, AndrewShinICP, PanMYLR15, XuXiChAAAI2015, BallasYPC15, YuWHYX15} proposed methods to model the temporal structure of videos as well as language.

The majority of these methods are inspired by sequence-to-sequence~\citep{s2sIlya} and attention~\citep{BahdanauCB14} models. Sequence learning was proposed to map the input sequence of a source language to a target language~\citep{s2sIlya}. Applying this method with an additional attention mechanism to the problem of translating a video to a description showed promising initial results, however, revealed additional challenges. First, modelling the video content with a fixed-length vector in order to map it to a language space is a more complex problem than mapping from a language to a language, given the complexity of visual content and the difference between the two modalities. Since not all frames in a video are equally salient for a short description, and an event can happen in multiple frames, it is important for a model to identify which frames are most salient.  Further, the models need additional work to be able to focus on points of interest within the video frames to select what to talk about.  Even a variable-length vector to represent a video using attention \citep{yao2015capgenvid} can have some problems. 

More specifically, current attention methods are local~\cite{YangYWSC16}, since the attention mechanism works in a sequential structure, and lack the ability to capture global structure. Moreover, combining a video and a description as a sequence-to-sequence problem motivates using some variant of a recurrent neural network (RNN) \citep{Hochreiter}: Given the limited capacity of a recurrent network to model very long sequences, memory networks~\citep{WestonCB14,SukhbaatarSWF15} have been introduced to help the RNN memorize sequences. However, one problem these memory networks suffer from is difficulty in training the model. The model proposed by \citet{WestonCB14} requires supervision at each layer, which makes training with backpropagation a challenging task. \citet{SukhbaatarSWF15} proposed a memory network that can be trained end-to-end, and the current work follows this research line to tackle the challenging problem of modeling vision and language memories for video description generation. 

%% file: Model.tex
\section{Learning to Attend and Memorize}
\begin{figure*}[t]
    \centering
    \includegraphics[width=0.62\textwidth]{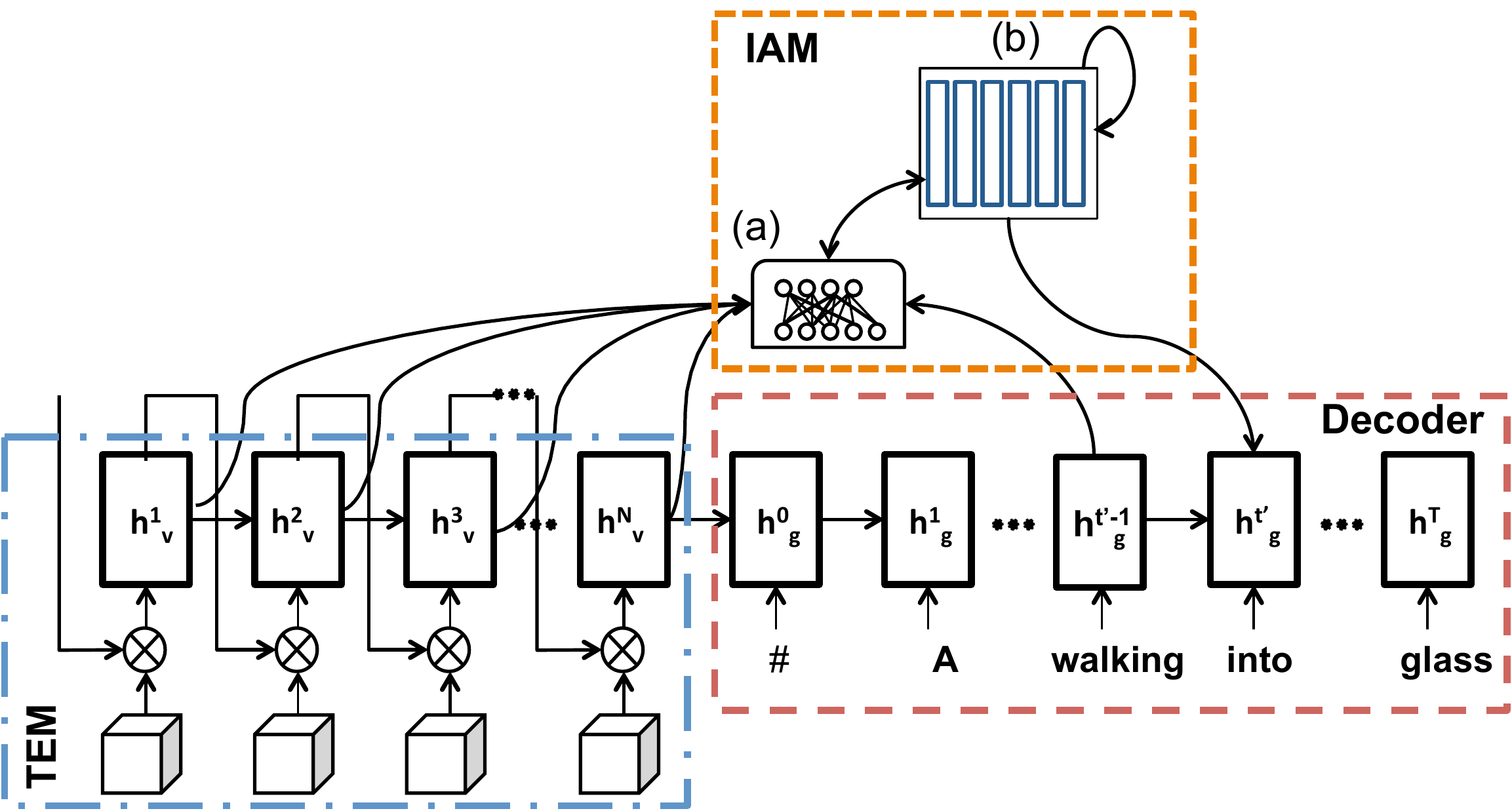}
    \caption{Our proposed architecture. Each component of our model is described in \ref{sec:TEM} through \ref{sec:Dec}.} 
    \label{fig:our_model}
\end{figure*}

A main challenge in video description is to find a mapping that can capture the connection between the video frames and the video description.  Sequence-to-sequence models, which work well at connecting input and output sequences in machine translation~\citep{s2sIlya}, do not perform as well for this task, as there is not the same direct alignment between a full video sequence and its summarizing description.

Our goal in the video description problem is to create an architecture that learns which moments to focus on in a video sequence in order to generate a summarizing natural language description.  The modelling challenges we set forth for the video description problem are: (1) Processing the temporal structure of the video; (2) Learning to attend to important parts of the video; and (3) Generating a description where each word is relevant to the video. At a high-level, this can be understood as having three primary parts: {\it When} moments in the video are particularly salient; {\it what} concepts to focus on; and {\it how} to talk about them.  We directly address these issues in an end-to-end network with three primary corresponding components (Figure \ref{fig:our_model}): A Temporal Model ({\sc tem}), An Iterative Attention/Memory Model ({\sc iam}), and a Decoder.  In summary:

\begin{itemize}
\item {\bf When:} Frames within the video sequence - The Temporal Model ({\sc tem}).
\item {\bf What:} Language-grounded concepts depicted in the video - The Iterative Attention/Memory mechanism ({\sc iam}).
\item {\bf How:} Words that fluently describe the {\it what} and {\it when} - The Decoder.
\end{itemize}


The Temporal Model is in place to capture the temporal structure of the video:  It functions as a {\it when} component.  The Iterative Attention/Memory is a main contribution of this work, functioning as a {\it what} component to remember relationships between words and video frames, and storing longer term memories.  The Decoder generates language, and functions as the {\it how} component to create the final description. 


To train the system end to end, we formulate the problem as sequence learning to maximize the probability of generating a correct description given a video:
\begin{equation}
\Theta^* = \underset{\Theta}{\arg\max}\sum_{(S, {f_1,\dots, f_N})} \log~p(S|f_1, \dots,f_N ;\mathbf{\Theta})
\end{equation}
where $S$ is the description, ${f_1, f_2,\dots,f_N}$ are the input video frames, and $\Theta$ is the model parameter vector. In the next sections, we will describe each component of the model, then explain the details of training and inference.  

\small \paragraph{\small Notational note:} Numbered equations use bold face to denote multi-dimensional learnable parameters, e.g., ${\mathbf{W^j_p}}$.  To distinguish the two different sets of time steps, one for video frames and one for words in the description, we use the notation $t$ for video and $t^\prime$ for language.  Throughout, the terms {\it description} and {\it caption} are used interchangeably.
\normalsize 
\subsection{Temporal Model ({\sc tem})}\label{sec:TEM} 

The first module we introduce encodes the temporal structure of the input video. A clear framework to use for this is a Recurrent Neural Network (RNN), which has been shown to be effectual in modelling the temporal structure of sequential data such as video \citep{BallasYPC15, SharmaKS15, venugopalan15iccv} and speech \citep{graves14}. In order to apply this in video sequences to generate a description, we seek to capture the fact that frame-to-frame temporal variation tends to be local \citep{BroxMalik2011} and critical in modeling motion~\citep{BallasYPC15}. 
Visual features extracted from the last fully connected layers of Convolutional Neural Networks (CNNs) have been shown to produce state-of-the-art results in image classification and recognition \citep{Simonyan14c, He_2016_CVPR}, and thus seem a good choice for modeling visual frames.  However, these features tend to discard low level information useful in modeling the motion in the video \citep{BallasYPC15}. 

To address these challenges, we implement an RNN we call the Temporal Model ({\sc tem}). 
At each time step of the {\sc tem}, a video frame encoding from a CNN 
serves as input. Rather than extracting video frame features from a fully connected layer 
of the pretrained CNN, we extract intermediate convolutional maps.

In detail, for a given video $X$  with $N$ frames $X = [X^1, X^2, \cdots, X^N]$,  $N$ convolutional maps of size $ R ^{L \times D}$ are extracted, where $L$ is the number of locations in the input frame and $D$ is the number of dimensions (See {\sc tem} in Figure \ref{fig:our_model}).
To enable the network to store the most important $L$ locations of each frame, 
we use a soft location attention mechanism, $f_{\mathbf{Latt}}$ \citep{BahdanauCB14, icml2015_xuc15, SharmaKS15}.  
We first use a softmax to compute $L$ probabilities that specify the importance of different parts in the frame, and this creates an input map for $f_{\mathbf{Latt}}$.

Formally, given a video frame at time $t$, $X^t \in R^{L \times D}$, the $f_{\mathbf{Latt}}$ mechanism is defined as follows:
\begin{align} \label{eq:Latt}
& {\rho^t_j} = \frac{ \exp( \mathbf{ W_p^j} h^{t-1}_v )}{\sum_{k=1}^L \exp(\mathbf{W_p^k} h^{t-1}_v )} \\
& f_{\mathbf{Latt}}({X^t, h^{t-1}_v};\mathbf{W_p}) = \sum_{j=1}^L {\rho_j^t} {X^t_{j}}
\end{align}
where $h^{t-1}_v \in R^K$ is the hidden state of the {\sc tem} at time $t$-1 with $K$ dimensions, and $W_p \in R^{L \times K}$. 
For each video frame time step, {\sc tem} learns a vector representation by applying location attention on the frame convolution map, conditioned on all previously seen frames:
\begin{align} \label{eq:temporal_Latt}
& {F^{t}} = f_{\mathbf{Latt}}({X^t, h^{t-1}_v};\mathbf{W_p}) \\
& {h^{t}_v} = f_\mathbf{v}({F^{t}},~{h^{t-1}_v};\mathbf{\Theta_v} )
\end{align}
where $f_\mathbf{v}$ can be an RNN/LSTM/GRU cell and {\bf $\Theta_v$} is the parameters of the $f_\mathbf{v}$. Due to the fact that vanilla RNNs have gradient vanishing and exploding problems~\citep{pascanu2013difficulty}, we use gradient clipping, and an LSTM with the following flow to handle potential vanishing gradients:
\begin{align*}
& {i^t} = \sigma(F^{t}  \mathbf{W_{x_i}} + {(h^{t-1}_v)}^T\mathbf{W_{h_i}}) \\
& {f^t} = \sigma(F^{t} \mathbf{W_{x_f}} + {(h^{t-1}_v)}^T\mathbf{W_{h_f}}) \\
& {o^t} = \sigma(F^{t} \mathbf{W_{x_o}} + {(h^{t-1}_v)}^T\mathbf{W_{h_o}}) \\
& {g^t} = {\tanh}(F^{t} \mathbf{W_{x_g}} + {(h^{t-1}_v)}^T\mathbf{W_{h_g}})    \\
& {c^t_v} =  {f^t} \odot {c^{t-1}_v} + {i^t} \odot {g^{t}} \\
& {h^t_v}  = {o_t}\odot {\tanh(c^t)}
\end{align*}
where $W_{h*} \in R^{K \times K}$, $W_{x*} \in R^{D \times K}$, and we define $\Theta_v = \{W_{h*},W_{x*}\}$.

\subsection{Iterative Attention/Memory ({\sc iam})}\label{sec:HAM} 

A main contribution of this work is a global view for the video description task: A memory-based attention mechanism that learns iterative attention relationships in an efficient sequence-to-sequence memory structure.
We refer to this as the Iterative Attention/Memory mechanism ({\sc iam}), and it aggregates information from previously generated words and all input frames. 

The {\sc iam} component is an iterative memorized attention between an input video and a description. More specifically, it learns a iterative attention structure for where to attend in a video given all previously generated words (from the Decoder), and previous states (from the {\sc tem}). This functions as a memory structure, remembering encoded versions of the video with corresponding language, and in turn, enabling the Decoder to access the full encoded video and previously generated words as it generates new words. 

This component addresses several key issues in generating a coherent video description.  In video description, a single word or phrase often describes action spanning multiple frames within the input video.  By employing the {\sc iam}, the model can effectively capture the relationship between a relatively short bit of language and an action that occurs over multiple frames. This also functions to directly address the problem of identifying which parts of the video are most relevant for description. 

The proposed Iterative Attention/Memory mechanism is formalized with an {\bf Attention} update and a {\bf Memory} update, detailed in Figure \ref{fig:HAM}.  Figure \ref{fig:our_model} illustrates where the {\sc iam} sits within the full model, with the Attention module shown in \ref{fig:our_model}a and the Memory module shown in \ref{fig:our_model}b.

\begin{figure}[t]
\begin{mdframed}
\begin{itemize}[leftmargin=.5em]
\item{Given:}
\small
\item[] \hspace{-1em} $N = \text{Number of frames in a given video}$ \vspace{-.5em}
\item[] \hspace{-1em} $T = \text{Number of words in description}$ \vspace{-.5em}
\item[] \hspace{-1em} $H_v = \text{Input video states}, [h_v^1, ..., h_v^N]$ \vspace{-.5em} 
\item[] \hspace{-1em} $H_g^{t^\prime-1} = \text{Decoder state~} h_g \text{~at time t-1, repeated N times}$ 
\item[] \hspace{-1em} $H_m^{t^\prime-1} = \text{Memory state~} h_m \text{~at time t-1, repeated N times}$ 
\item[] \hspace{-1em} $W_v, W_g \in R^{ K\times K}$ \vspace{-.5em}
\item[] \hspace{-1em} $W_m \in R^{M\times K}$ \vspace{-.5em}
\item[] \hspace{-1em} $u \in R^{K}$ \vspace{-.5em} 
\item[] \hspace{-1em} $\alpha$ = Probability over all N frames \vspace{-1em}
\item[] \hspace{-1em} $\Theta_a = \{W_v, W_g, W_m, u\}$
\normalsize
\item{Attention update [${\hat{F}(\mathbf{\Theta_a})}$]:} \vspace{-.25em}
\small
\begin{align}\label{eq:att_update}
& \hspace{-1.5em}  {Q_A} = \tanh({H_v}\mathbf{W_v}+{H_g^{t^\prime -1 }}\mathbf{W_g}+{H_m^{t^\prime -1 }}\mathbf{W_m}) \\
&\hspace{-1.5em} \mathbf{\alpha}_{t^\prime} = \mathrm{softmax}( {Q_{A}} \mathbf{u} )\\
&\hspace{-1.5em} {\hat{F}} =  {H_v^T}\alpha_{t^\prime}
\end{align}
\normalsize
\item{Memory update:}  \vspace{-.25em}
\small
\begin{align}\label{eq:mem_update}
& \hspace{-1.5em}\vspace{-10em} {h_m^{{t^\prime}}} = f_m({h_m^{t^\prime -1 }},~{\hat{F}};\mathbf{\Theta_m)}~~~~~~~~~~~~~~~~~~~~~~~~~~~~~~~~~~~~~~~
\end{align}
\end{itemize}
\end{mdframed}
\caption{Iterative Attention and Memory ({\sc iam}) is formulated as an Attention update and a Memory update.}
\label{fig:HAM}
\end{figure}

As formalized in Figure \ref{fig:HAM}, the {\it Attention} update $\hat{F}(\Theta_a)$ computes the set of probabilities in a given time step for attention within the input video states, the memory state, and the decoder state. 
The {\it Memory} update stores what has been attended to and described. This serves as the memorization component, combining the previous memory with the current iterative attention $\hat{F}$. 
We use an LSTM $f_m$ with the equations described above to enable the network to learn multi-layer attention over the input video and its corresponding language. The output of this function is then used as input to the Decoder. 

\subsection{Decoder}\label{sec:Dec}
In order to generate a new word conditioned on all previous words and {\sc iam} states, a recurrent structure is modelled as follows:
\begin{align} \label{eq:Dec}
&h^{{t^\prime }}_g = f_g(s^{t^\prime}, ~h_m^{t^\prime},~h^{{t^\prime-1}}_g; \mathbf{\Theta_g}) \\
&\hat{s}^{t^\prime} = \mathrm{softmax}((h^{{t^\prime}}_g)^T\mathbf{W_e})
\end{align}
where $h^{t^\prime}_g \in R^K$, $s^{t^\prime}$ is a word vector at time ${t^\prime}$, $W_e\in R^{K \times |V|}$, and $|V|$ is the vocabulary size. In addition, $\hat{s}_{t^\prime}$ assigns a probability to each word in the language. $f_{g}$ is an LSTM where $s^{t^\prime}$ and $h_m^{t^\prime}$ are inputs and $h^{{t^\prime}}_g$ is the recurrent state.
\subsection{Training and Optimization }\label{sec:training}
The goal in our network is to predict the next word given all previously seen words and an input video. In order to optimize our network parameters $\Theta = \{W_p, \Theta_v, \Theta_a, \Theta_m,  \Theta_g, W_e \} $, we minimize a negative log likelihood loss function:
\begin{equation}\label{eq:NLL}
L (S, X; \mathbf{\Theta}) = -\sum_{t^\prime}^T \sum_{i}^{|V|} s_i^{t^\prime}\log (\hat{s}_i^{t^\prime}) + \lambda\parallel\Theta \parallel_2^2 
\end{equation}
where $|V|$ is the vocabulary size. We fully train our network in an \textit{end-to-end} fashion using first-order stochastic gradient-based optimization method with an adaptive learning rate. More specifically, in order to optimize our network parameters, we use Adam~\citep{KingmaB14} with learning rate $2\times 10^{-5}$ and set $\beta_1$, $\beta_2$ to $0.8$ and $0.999$, respectively. During training, we use a batch size of $16$. The source code for this paper is available on \url{https://github.com/rasoolfa/videocap}.

%% file: Experiments.tex
\section{Experiments}
\paragraph*{Dataset} We evaluate the model on the \textit{Charades}~\citep{sigurdsson2016hollywood} dataset and the {\it Microsoft Video Description Corpus (MSVD)}~\citep{chencl11}. Charades contains $9,848$ videos (in total) and provides $27,847$\footnote{Only $16087$ out of $27,847$ are used as descriptions for our evaluation since the $27,847$ refers to script of the video as well as descriptions.} video descriptions. We follow the same train/test splits as \citet{sigurdsson2016hollywood}, with $7569$ train, $1,863$ test, and $400$ validation. A main difference between this dataset and others is that it uses a ``Hollywood in Homes'' approach to data collection, where ``actors'' are crowdsourced to act out different actions.  This yields a diverse set of videos, with each containing a specific action. 

MSVD is a set of YouTube videos annotated by workers on Mechanical Turk,\footnote{\url{https://www.mturk.com/mturk/welcome}} who were asked to pick a video clips representing an activity. In this dataset, each clip is annotated by multiple workers with a single sentence. The dataset contains $1,970$ videos and about $80,000$ descriptions, where $1,200$ of the videos are training data, $670$ test, and the rest ($100$ videos) for validation. In order for the results to be comparable to other approaches, we follow the \textit{\textbf{exact}} training/validation/test splits provided by \citet{venugopalannaacl15}.

\paragraph*{Evaluation metrics}
We report results on the video description generation task. In order to evaluate descriptions generated by our model, we use model-free automatic evaluation metrics. We adopt \textsc{meteor}, \textsc{bleu-n}, and \textsc{cide}r metrics available from the Microsoft COCO Caption Evaluation code\footnote{\url{https://github.com/tylin/coco-caption}} to score the system.

\paragraph*{Video and Caption preprocessing}
We preprocess the captions for both datasets using the Natural Language Toolkit (NLTK)\footnote{\url{http://www.nltk.org/}} and clip each description up to $30$ words,  since the majority have less. 
We extract sample frames from each video and pass each frame through VGGnet~\citep{Simonyan14c} without fine-tuning. For the experiments in this paper, we use the feature maps from \textit{conv5\_3} layer after applying \textit{ReLU}. The feature map in this layer is $14\times 14 \times 512$. Our {\sc tem} component operates on the flattened $196\times 512$ of this feature cubes. For the ablation studies, features from the fully connected layer with $4096$ dimensions are used as well.

\paragraph*{Hyper-parameter optimization}
We use random search~\citep{Bergstra2012} on the validation set to select hyper-parameters on both datasets. The word-embedding size, hidden layer size (for both the {\sc tem} and the Decoder), and memory size of the best model on Charades are: $237$, $1316$, and $437$, respectively. These values are  $402$, $1479$, and $797$ for the model on the MSVD dataset. A stack of two LSTMs are used in the Decoder and {\sc tem}. The number of frame samples is a hyperparameter which is selected among $4$, $8$, $16$, $40$ on the validation set.  \textsc{att + No {\sc tem}} and \textsc{No iam + {\sc tem}} get the best results on the validation set with $40$ frames, and we use this as the number of frames for all models in the ablation study.

\subsection{Video Caption Generation}
We first present an ablation analysis to elucidate the contribution of the different
components of our proposed model. Then, we compare the overall performance of our model to other recent models.
\subsection*{Ablation Analysis}
Ablation results are shown in Table~\ref{tab:abl}, evaluating on the MSVD test set. 
The first (\textsc{Att + No tem}) corresponds to a simpler version of our model in which we remove the {\sc tem} component and instead pass each frame of the video through a CNN, extracting features from the last fully-connected hidden layer. 
In addition, we replace our {\sc iam} with a simpler version where the model only memorizes the current step instead of all previous steps. In the next variation (\textsc{Att + tem}), it is same as the first one except we use {\sc tem} instead of fully connected CNN features. In the next ablation (\textsc{No iam + tem}), we remove the {\sc iam} component from our model and keep the rest of the model as-is. In the next variation (\textsc{iam + No tem}), we remove the {\sc tem} and calculate features for each frame, similar to \textsc{Att + No tem}. Finally, the last row in the table is our proposed model (\textsc{iam + tem}) with all its components.

The {\sc iam} plays a significant role in the proposed model, and removing it causes a large drop in performance, as measured by both {\sc bleu} and {\sc meteor}. On the other hand, removing the {\sc tem} by itself does not drop performance as much as dropping the {\sc iam}. Putting the two together, they complement one another to result in overall better performance for {\sc meteor}. However, further development on the {\sc tem} component in future work is warranted. In the \textsc{No iam + tem} condition, an entire video must be represented with a fixed-length vector, which may contribute to the lower performance~\citep{BahdanauCB14}. This is in contrast to the other models, which apply single layer attention or {\sc iam} to search relevant parts of the video aligned with the description. 

\begin{table*}[t]
\small
\centering
    \begin{tabularx}{.75\linewidth}{l|llllll}
      \toprule 
      Method & {\sc meteor} & {\sc bleu-1} &{\sc bleu-2} & {\sc bleu-3} & {\sc bleu-4} & {\sc cide}r\\ 
      \midrule
      \textsc{Att + No tem} & $31.20 $ & $77.90 $ & $65.10 $  & $ 55.30$ & $44.90 $ &  $\mathbf{63.90}$ \\
      \textsc{Att + tem} & $31.00 $ & $79.00 $ & $66.50 $  & $ 56.30$ & $45.50 $ &  $61.00$ \\
      \textsc{No iam + tem} & $30.50$ & $78.10$ & $65.20$  & $55.10$ & $44.60$ &  $60.50$ \\
      \textsc{iam + No tem} & $31.00 $ & $ 78.70$ & $66.90$  & $\mathbf{57.40} $ & $\mathbf{47.00} $ &  $62.10 $ \\
      \textsc{iam + tem [40f]}     & $31.70$ & $79.00$ & $66.20$  & $56.0$ & $45.60$ & $62.20 $ \\
      \textsc{iam + tem [8f]}     & $\mathbf{31.80}$ & $\mathbf{79.40}$ & $\mathbf{67.10}$  & $56.80$ & $46.10$ & $62.70 $ \\

    \end{tabularx} 
        \caption{Ablation of proposed model with and without the {\sc iam} component on the MSVD test set.\label{tab:abl}}
    
\end{table*}

\subsection*{Performance Comparison}

\begin{table*}[t]
\small
\centering
    \begin{tabularx}{.875\textwidth}{lllllll}
      \toprule 
      Method & {\sc meteor} & {\sc bleu-1} &{\sc bleu-2} & {\sc bleu-3} & {\sc bleu-4} & {\sc cide}r\\ 
            \midrule
      \small{\citet{venugopalannaacl15}}  & $27.7$ & $-$ & $-$  & $-$ & $-$  & $-$\\
      \small{\citet{venugopalan15iccv}}  & $29.2$ & $-$ & $-$  & $-$ & $-$& $-$ \\
      \small{\citet{PanMYLR15}}  & $29.5$ & $74.9$ & $60.9$  & $50.6$ & $40.2$ & $-$\\
      \small{\citet{YuWHYX15}}  & $31.10 $ & $77.30 $ & $64.50 $  & $54.60$ & $44.30$ & $-$\\
      \small{\citet{pan2016hierarchical}} & $\mathbf{\underline{33.10}} $ & $79.20 $ & $66.30 $  & $55.10 $ & $43.80$ & $-$\\
      \textbf{Our Model}        & $31.80$ & $\mathbf{\underline{79.40}}$ & $\mathbf{\underline{67.10}}$  & $\mathbf{\underline{56.80}}$ & $\mathbf{\underline{46.10}}$ & $\mathbf{\underline{62.70}} $ \\
      \midrule
       \small{\citet{yao2015capgenvid} + C3D} &  $29.60$ & $-$ & $-$  & $-$ & $41.92 $ & $51.67$\\
      \small{\citet{venugopalan15iccv} + Flow}  & $29.8$ & $-$ & $-$  & $-$ & $-$ &  $-$\\
      \small{\citet{BallasYPC15}} + FT  & $ 30.75$ & $-$ & $-$  & $-$ & $49.0$ &  $59.37$\\
      \small{\citet{PanMYLR15} + C3D}  & $31.0 $ & $78.80 $ & $66.0 $  & $55.4 $ & $45.3$ &  $-$\\
      \small{\citet{YuWHYX15} + C3D} & $ 32.60$ & $81.50 $ & $70.40 $  & $60.4$ & $49.90$ &  $- $\\
    \end{tabularx} 
        \caption{Video captioning evaluation on MSVD ($670$ videos). \label{tab:main_res}}
\end{table*}

\begin{table}[t]
\small
\centering
    \begin{tabularx}{\columnwidth}{@{}l@{\hspace{.15em}}|@{\hspace{.25em}}l@{\hspace{.4em}}l@{\hspace{.25em}}l@{\hspace{.4em}}l@{\hspace{.4em}}l@{\hspace{.4em}}l@{}}
      \toprule 
      Method & M & B@1 & B@2 & B@3 & B@4 & C\\ 
      \midrule
      \small{Human}  & \multirow{2}{*}{$24$} & \multirow{2}{*}{$62$} & \multirow{2}{*}{$43$}  & \multirow{2}{*}{$29$} & \multirow{2}{*}{$20$} &  \multirow{2}{*}{$53$}\\
      \citep{sigurdsson2016hollywood} \\
      \midrule
      \midrule
      \small{\citet{sigurdsson2016hollywood}}  & $16$ & $49$ & $30$  & $18$ & $11$ &  $14$\\
      \textbf{Our Model}       & $\mathbf{17.6} $ & $\mathbf{50}$ & $\mathbf{31.1} $  & $\mathbf{18.8} $ & $\mathbf{11.5} $ &  $\mathbf{16.7} $\\
    \end{tabularx} 
        \caption{Video captioning evaluation on Charades ($1863$ videos). M={\sc Meteor}, B={\sc bleu}, C={\sc cide}r. \citet{sigurdsson2016hollywood} results use the \citet{venugopalan15iccv} model. \label{tab:main_res_HW}}
\end{table}

To extensively evaluate the proposed model, we compare with state-of-the-art models and baselines for the video caption generation task on the MSVD dataset. In this experiment, we use $8$ frames per video as the inputs to the {\sc tem} module. As shown in Table \ref{tab:main_res},\footnote{The symbol $-$ indicates that the score was not reported by the corresponding paper.  The horizontal line in Table \ref{tab:main_res} separates models that do/do not use external features for the video representation.} our proposed model achieves state-of-the-art scores in {\sc bleu}-4, and outperforms almost all systems on {\sc meteor}.  The closest-scoring comparison system, from \citet{pan2016hierarchical}, shows a trade-off between {\sc meteor} and {\sc bleu}:  {\sc bleu} prefers descriptions with short-distance fluency and high lexical overlap with the observed descriptions, while {\sc meteor} permits less direct overlap and longer descriptions. A detailed study of the generated descriptions between the two systems would be needed to better understand these differences.

The improvement over previous work is particularly noteworthy because we do not use external features for the video, such as Optical Flow \citep{Bro04a} (denoted Flow), 3-Dimensional Convolutional Network features~\citep{C3DTan} (denoted C3D), or fine-tuned CNN features (denoted FT), which further enhances aspects such as action recognition by leveraging an external dataset such as UCF-101. The only system using external features that outperforms the model proposed here is from \citet{YuWHYX15}, who uses a slightly different version of the same dataset\footnote{\citet{YuWHYX15} uses the MSVD dataset reported in \cite{Guadarrama2013}, which has different preprocessing.} along with C3D features for a large improvement in results (compare Table~\ref{tab:main_res} rows 4 and 11); future work may explore the utility of external visual features for this work.  Here, we demonstrate that the proposed architecture maps visual space to language space with improved performance over previous work, before addition of further resources.

We additionally report results on the Charades dataset \cite{sigurdsson2016hollywood}, which is challenging to train on because there are only a few ($\approx2$) captions per video. In this experiment, we use $16$ frames per video as the input to the {\sc tem} module. As shown in Table~\ref{tab:main_res_HW}, our method achieves a $10\%$ relative improvement over the \citet{venugopalan15iccv} model reported by \citet{sigurdsson2016hollywood}. It is worth noting that humans reach a {\sc meteor} score of $24$ and a {\sc bleu-4} score of $20$, illustrating the low upper bound in this task.\footnote{For comparison, the upper bound {\sc bleu} score in machine translation for English to French is above 30.}

\begin{figure*}[t]
    \centering
    \begin{tabular}{ll}
    \includegraphics[trim=0cm 2.15cm 0cm 3cm, clip=true, width=\columnwidth]{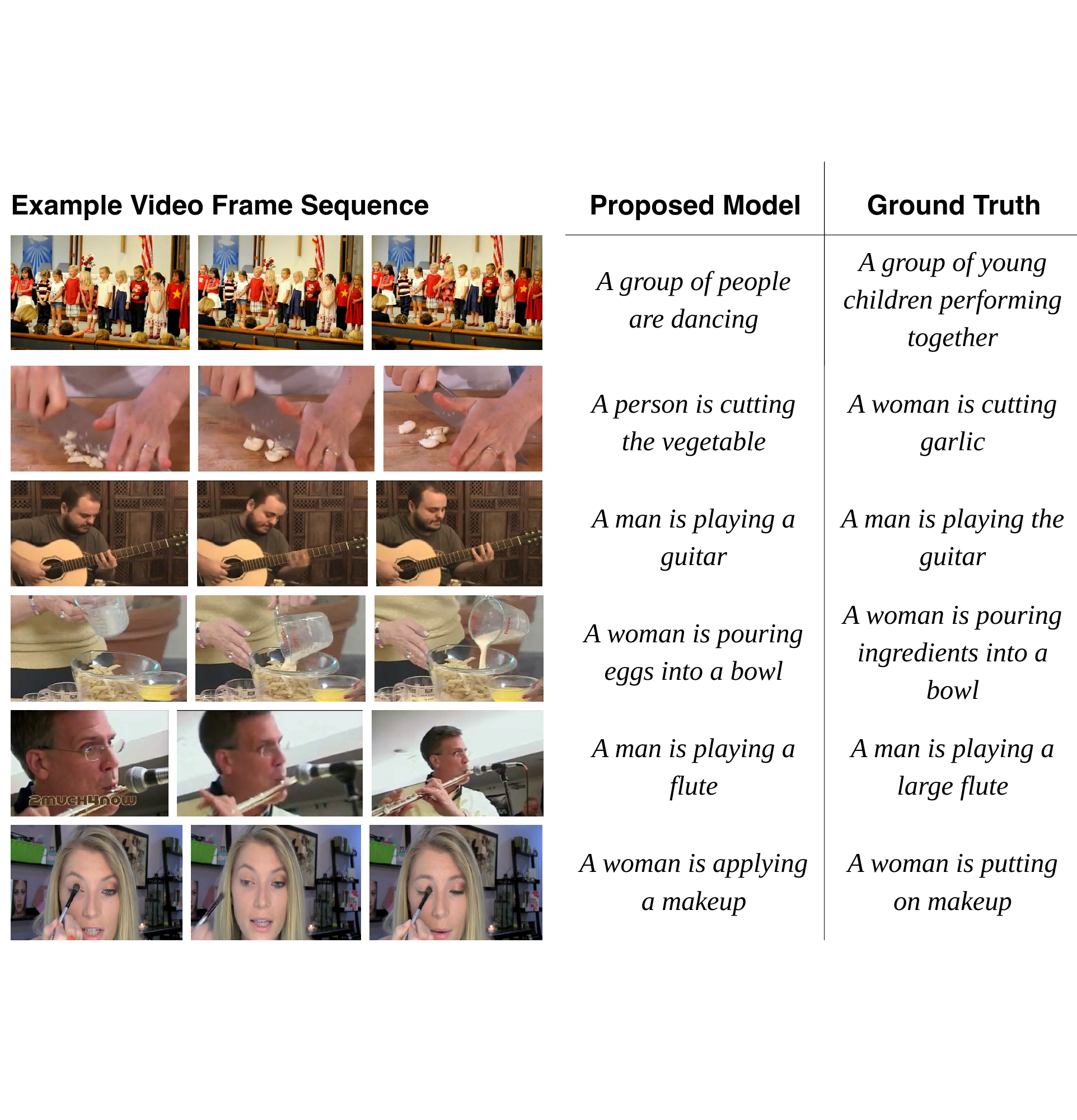}&
   \includegraphics[trim=0cm 3cm 0cm 1cm, clip=true, width=\columnwidth]{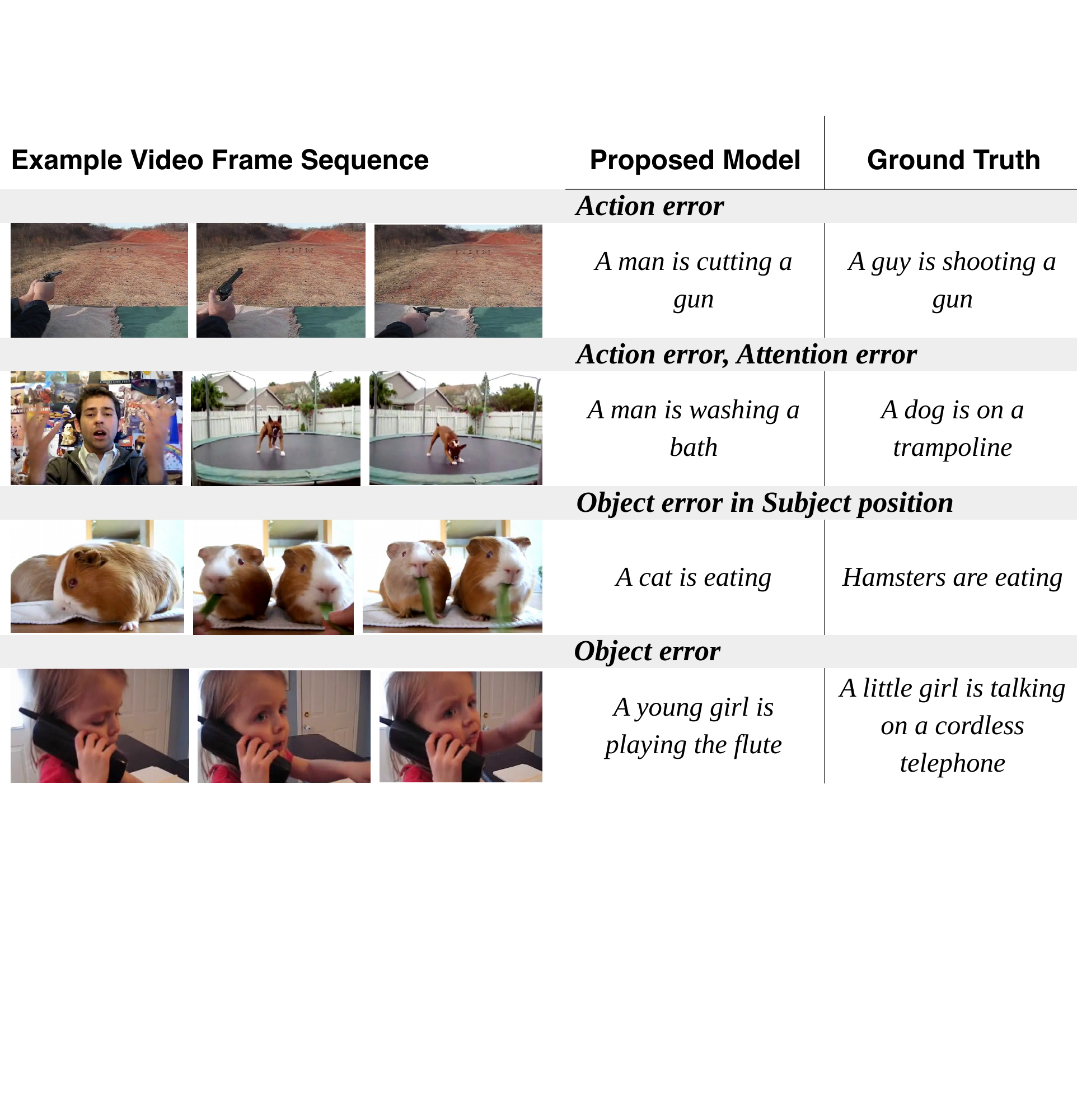}
    \end{tabular}
    \caption{Example captions generated by our model on MSVD test videos.}
    \label{fig:caps_samples}
\end{figure*}

\subsection*{Results Discussion}
We show some example descriptions generated by our system in Figure \ref{fig:caps_samples}. The model generates mostly correct descriptions, with naturalistic variation from the ground truth.  Errors illustrate a preference to describe items that have a higher likelihood of being mentioned, even if they appear in less of the frames.  For example, in the ``a dog is on a trampoline" video, our model focuses on the man, who appears in only a few frames, and generates the incorrect description ``a man is washing a bath".  The errors, alongside the ablation study shown in Table \ref{tab:abl}, suggest that the {\sc tem} module in particular may be further improved by focusing on how frames in the video sequence are captured and passed to the {\sc iam} module.

%% file: Conclusion.tex
\section{Conclusion}
We introduce a general framework for an memory-based sequence learning model, trained end-to-end.  We apply this framework to the task of describing an input video with a natural language description. Our model utilizes a deep learning architecture that represents video with an explicit model of the video's temporal structure, and jointly models the video description and the temporal video sequence.  This effectively connects the visual video space and the language description space.

A memory-based attention mechanism helps guide where to attend and what to reason about as the description is generated. This allows the model to not only reason efficiently about local attention, but also to consider the full sequence of video frames during the generation of each word. Our experiments confirm that the memory components in our architecture, most notably from the {\sc iam} module, play a significant role in improving the performance of the entire network. 

Future work should raim to refine the temporal video frame model, {\sc tem}, and explore how to improve performance on capturing the ideal frames for each description.